\documentclass{article}

\usepackage{amsmath, amsthm, amssymb}
\usepackage{graphicx}
\usepackage{verbatim}
\usepackage{natbib}
\usepackage{caption}
\usepackage{subcaption}
\usepackage{fancyvrb}
\usepackage{enumerate}
\usepackage{relsize}
\usepackage{float}
\usepackage{graphicx}
\usepackage[english]{babel}
\usepackage{algorithm}
\usepackage[noend]{algpseudocode}
\usepackage{tikz}
\usetikzlibrary{fit,positioning}

\usepackage{hyperref}
\usepackage[margin=1.5in]{geometry}
\hypersetup{colorlinks,citecolor=blue,urlcolor=blue,linkcolor=blue}

\usepackage{stefan_tex}
\graphicspath{{./figures/}}

\usepackage{lscape}

\usepackage{multicol}

\theoremstyle{plain}
\newtheorem{prop}{Proposition}

\newtheorem{lemm}[prop]{Lemma}

\theoremstyle{definition}

\theoremstyle{remark}

\author{Jonathan Johannemann \\ \texttt{jonjoh@stanford.edu}
\and Vitor Hadad \\ \texttt{vitorh@stanford.edu}
\and Susan Athey \\ \texttt{athey@stanford.edu}
\and Stefan Wager \\ \texttt{swager@stanford.edu}}
\date{Stanford University}
\title{Sufficient Representations for Categorical Variables}

\begin{document}

\maketitle

\begin{abstract}
    \noindent Many learning algorithms require categorical data to be transformed into real vectors before it can be used as input. Often, categorical variables are encoded as \emph{one-hot} or \emph{dummy} vectors. However, this mode of representation can be wasteful since it adds many low-signal regressors, especially when the number of unique categories is large.  In this paper, we investigate simple alternative solutions for universally consistent estimators that rely on lower-dimensional real-valued representations of categorical variables that are \emph{sufficient} in the sense that no predictive information is lost. We then compare preexisting and proposed methods on simulated and observational datasets.
\end{abstract}

\section{Introduction}

Many regression problems involve data collected from a number groups that may be statistically relevant.
For example, in a medical setting, we may want to model health outcomes using data on patients from
several hospitals and acknowledge that different hospitals may have idiosyncratic effects on patients
that are not explained by other covariates. Similar considerations arise when working with data on
students from different schools, voters from different zip-codes, employees at different firms, etc.

One of the most wide-spread approaches to this problem is via fixed effect modeling,
as follows. Suppose that we observe $n$ samples $\p{X_i, \, G_i, \, Y_i}$
for $i = 1, \, ..., \, n$, where $X_i \in \RR^p$ is a set of subject-specific covariates, $G_i \in \gcal$ is a categorial variable that
records group membership and $Y_i \in \RR$ is the response of interest, and want to estimate
\begin{equation}
\label{eq:mu}
\mu(x, \, g) = \EE{Y_i \cond X_i = x, \, G_i = g}.
\end{equation}
Then, the simple fixed effects approach starts by positing a model
\begin{equation}
\label{eq:FE}
\mu(x, \, g) = \alpha_g + x\beta,
\end{equation}
and then estimating the coefficients $\beta$ and $\alpha_g$ via ordinary least squares regression.
More sophisticated extensions of this approach may involve considering non-linear transformations
of $x$, interactions between group membership and the covariates $x$, and/or regularization
\citep{angrist2008mostly,diggle2002analysis,wooldridge2010econometric}.

Fixed effects modeling, however, does not always perform well with complex non-linear signals
or when the number of groups $\abs{\gcal}$ is large. The model \eqref{eq:FE} is quite rigid
and may not be able to represent rich signals while, at the same time, the large number of $\alpha_g$
parameters in the model \eqref{eq:FE} may result in problems for statistical inference \citep{neyman1948consistent}.
In other words, the model \eqref{eq:FE} may have too many parameters to be stable all while lacking the
degrees of freedom to fit the signal well.

The goal of this paper is to develop a more parsimonious approach for representing group membership to
avoid the above problems. Specifically, we seek a mapping $\psi$ that embeds group membership $G_i$
into a $k$-dimensional space without losing any predictive information, i.e.,
\begin{equation}
\label{eq:repr}
\psi: \gcal \rightarrow \RR^k, \ \ \mu(x, \, g) = f(x, \, \psi(g)),
\end{equation}
such that $k$ is small (in particular, $k \ll \abs{\gcal}$) and the function $f(\cdot, \, \cdot)$ is still easy to learn.
Given such a mapping, the problem \eqref{eq:mu} becomes a routine regression problem with
$(p+k)$-dimensional real-valued features $(X_i, \, \psi(G_i))$, and we can use out-of-the-box statistical learning
software on it.

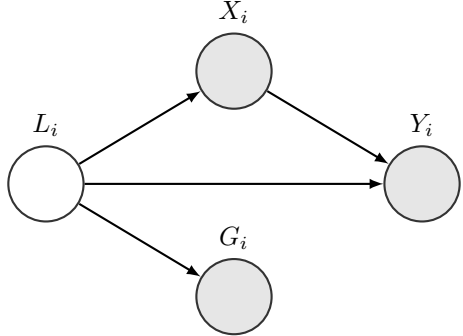
\begin{figure}[t]
\centering
\begin{tikzpicture}
\tikzstyle{main}=[circle, minimum size = 10mm, thick, draw =black!80, node distance = 14mm]
\tikzstyle{connect}=[-latex, thick]
\tikzstyle{box}=[rectangle, draw=black!100]
  \node[main] (L) [xshift=-2.5cm,label=$L_i$] { };
  \node[main, fill=black!10] (X) [yshift=1.5cm,label=$X_i$] {};
  \node[main, fill = black!10] (G) [yshift=-1.5cm,label=$G_i$] { };
  \node[main, fill=black!10] (Y) [xshift=2.5cm,label=$Y_i$] { };
  \path
        (L) edge [connect] (X)
        (L) edge [connect] (G)
        (X) edge [connect] (Y)
        (L) edge [connect] (Y);
\end{tikzpicture}
\caption{Causal graph depicting the key assumption that $Y_i$ and $X_i$ are independent of group
membership $G_i$ conditionally on latent state $L_i$. The grayed-out nodes are observed.}
\label{fig:graph_simple}
\end{figure}

In order to obtain a useful representation of group membership $G_i$, we of course need to assume
something about the relationship between $G_i$ and the target outcome $Y_i$. The core assumption
made in this paper is what we call the \emph{sufficient latent state assumption} depicted in Figure
\ref{fig:graph_simple}: group membership $G_i$ has no direct causal effect on $Y_i$, but may be
associated with latent variables $L_i$ that do have a direct effect on $Y_i$.
For example, in the case of patients spread across hospitals, we assume that hospitals themselves do not
directly \emph{cause} health outcomes $Y_i$; however, hospitals may still be \emph{predictive} of
$Y_i$ through their association with latent causal variables. Patients may have unobserved
characteristics, e.g., severity of disease or socioeconomic resources, that both affect $Y_i$ and lead the
patient to self-select into different hospitals. Our main result is that, under this sufficient latent state
assumption, practical representations of the form \eqref{eq:repr} exist and can be learned from data.

One important application area where regression problems of the form \eqref{eq:mu} are common arises in causal inference,
when we want to measure the effect of a treatment on an outcome of interest while controlling
for measured confounders \citep{imbens2015causal,rosenbaum1983central}. In such applications, it is
common to have a mix of continuous confounders (e.g., age, income, etc.) and high-cardinality categorical
ones (e.g., zip code, hospital, etc.). Given this type of data, modern approaches to causal inference start
by solving a series of non-parametric regression problems of the form \eqref{eq:mu} and then synthetize
them into a doubly robust treatment effect estimator; see \citet{athey2016approximate}, \citet{chernozhukov2016double},
\citet{robins1994estimation}, \citet{van2011targeted} and references therein. Software packages built around
this idea include \texttt{tmle} \citep{tmle} and \texttt{grf} \citep{athey2018generalized} for \texttt{R} \citep{CRAN}.
As discussed further in Section \ref{sec:DR}, we expect our approach to predictive modeling with high-cardinality
categorical variables to be particularly useful in the context of doubly robust treatment effect estimation.

Our paper is structured as follows. We begin by reviewing similar problem settings in the fixed effects literature and the drawbacks of using existing methods in \ref{subsec:related_work}. In Section 2, we introduce the primary lemma which seeks to describe the true information we wish to extract from categorical variables. In Section 3, we expand on lemma 1 to develop methods that utilize this insight. In Sections 4 and 5, we run simulated and observational experiments with our proposed methods and follow up with discussion on how realized performance compared to our expectations.

\subsection{Related Work}
\label{subsec:related_work}

The principle of representing high-cardinality categorical variables as real-valued vectors has played an important role in many different areas. For example, in econometrics the categories often represent groups the individual observations belong to (such as classroom cohorts or a region of the country) or are identified by (such as the individual identifier in longitudinal data).

Our work was originally motivated by a result of \citet{arkhangelsky2018role}, who showed that in many settings
where we want to control for a mix of categorical and continuous confounders $X_i$ and $G_i$ respectively, we don't need to explicitly control
for the categorical variable---and it's enough to control for the distribution of the continuous confounders across
samples in each category. Furthermore, if the distribution $X_i \cond G_i$ follows an exponential family, it's enough
to control for $\mu_{G_i}$, where $\mu_g = \EE{X_i \cond G_i = g}$. In a sense, the results of this paper can be
seen as an application of the ideas of \citet{arkhangelsky2018role} to the problem of generic predictive modeling.

In natural language processing, it is common
to start more complex analyses with a pre-processing step that represent words as vectors that capture
the way in which those words are used in context \citep{mikolov2013efficient,pennington2014glove}.
Meanwhile, \citet{cerda2018similarity} recently consider a related problem of representing ``dirty'' categorical
variables that might arise if, e.g., several categorical levels are just misspellings of each other, and
propose using a low-dimensional embedding that exploits lexicographic similarity (i.e., factors with
similar spellings are arranged close to each other). In this paper, we use information in the $X_i$,
rather than lexicographic information, to construct an embedding; however, the high-level conclusion
that we can achieve meaningful gains by using auxiliary information to embed categorical variables in
a low-rank space remains. Also, \citet{rahimi2008random} proposes randomly projecting a one-hot
representation of the categorical variables into $\RR^k$ for relatively small $k$.

Meanwhile, in the literature on panel data analysis, the approach we propose here is perhaps most closely related to the one presented in \citet{bonhomme2015grouped} where individual time series belong to discrete clusters and we have only one fixed effect per cluster (rather than one per time series). \citet{bonhomme2015grouped} then fit this model via a $k$-means like algorithm that alternates clustering and estimation with per-cluster fixed
effects. Our approach is not directly comparable to either of these methods, as we do not focus on textual data, and do not assume that the latent state $L_i$ can be consistently estimated (in contrast, \citet{bonhomme2015grouped} assume that they have access to long enough time series that their clustering step is consistent which, in our setting, would be equivalent to assuming that $L_i$ can be recovered).

In the broader machine learning literature, the discussion of how to best account for group membership $G_i$ in a non-parametric regression
has focused on different ways to encode $G_i$ that can be given as an input to
statistical software. One simple way to do so is via one-hot encoding:
\smash{$\iota: \gcal \rightarrow \{0, \, 1\}^M$} such that the $j$-th entry of \smash{$\iota(g)$}
is 1 if and only if $g$ corresponds to the $j$-th element in $\gcal$, and where $M := {\abs{\gcal}}$. Note that linear regression on
one-hot encoded features $(X_i, \, \iota(G_i))$ exactly recovers the standard fixed effects model
\eqref{eq:FE}.

As discussed above, however, one-hot encoding may lead to undesirably high-dimensional problems when $M := \abs{\gcal}$ is large.\footnote{Another slightly more subtle difficulty is that when the categorical variable has many levels, the individual features $\iota(G_i)_j$ become very sparse (i.e., they are usually
0 and only very rarely 1). Many approaches to statistical learning work better with features whose
variance roughly captures their range than with such spiky features.} In the  Appendix (\ref{app:encodings}),
we present multiple encoding methods that similarly project the categories onto $\mathbb{R}^{M}$.
These methods do not utilize information from the covariates $X_i$ or response $Y_i$ and suffer from the same pitfalls that
come with high dimensional representation of the observed groups $\gcal$. The primary difference for
these methods are the user's interpretation of the encoded variables which are commonly constructed as the comparison of
 the mean effect of a subset $\gcal' \in \gcal$ relative to the mean effect of the set $\gcal \backslash \gcal'$ or one of its subsets.

The problem of fixed effects is especially challenging with sparsity-seeking methods such as the
lasso \citep{hastie2015statistical} or decision trees \citep{breiman1984classification}, and related
ensemble methods such as random forests \citep{breiman2001random} or gradient-boosted
trees \citep{friedman2001greedy}. Sparsity-seeking methods will set the contribution of features
to zero unless there is strong evidence that those features matter for prediction, and it is difficult
for rare levels of $G_i$ to produce sufficient evidence to get a non-zero contribution to the model
via their one-hot features. The end result is that sparsity seeking methods may largely ignore high-cardinality
one-hot encoded factors.

Another prevalent way of working with categorical variables and decision trees
is to consider full factorial splits that allow for arbitrary grouping of the levels of the categorical
variable. For a variable with $M := \abs{\gcal}$ levels, this allows
for \smash{$2^{M - 1} - 1$} potential splits. \citet{breiman1984classification} showed that
we can optimize over this exponential set of potential splits in time that scales linearly in $M$;
however, from a statistical point of view, such factorial splits are prone to very strong overfitting
when the number of levels is large.

\section{Representing Groups with Sufficient Latent State}

Our \emph{sufficient latent state} assumption presented in the introduction and depicted in the causal graph \ref{fig:graph_simple} implies that the distribution of the outcome $Y_{i}$ only depends on the observable factor $G_{i}$ through some unobservable latent variable $L_{i}$. In other words, if we knew the value of $L_{i}$, then also knowing $G_{i}$ would give us no additional information about the outcome. For a simple example, one may posit that a patient's underlying health status ($L_{i} \in \{\text{good}, \text{poor}\}$) may simultaneously determine to which hospital they are admitted ($G_{i}$), what symptoms ($X_i$) they exhibit, and what health outcomes outcomes ($Y_i$) they attain. Conditioned on the underlying health status, the hospital cannot provide any additional information about any of the other variables. Conversely, learning their hospital is only helpful inasmuch it allows us to infer something about their health status.

The following lemma states that we can characterize \emph{how} the information about the categorical variable $G_{i}$ enters the model: the conditional expectation function of the outcome depends only on the \emph{conditional probabilities of the latent variable given the observable category}. This fact will be crucial when deriving the representation methods in future sections.

\begin{lemm}
\label{lemm:repr}
Suppose that the latent state $L_i$ is discrete with $k$ possible levels, and that the probabilistic structure required by the sufficient latent state assumption (Figure \ref{fig:graph_simple}) holds. Then,
\begin{equation}
\label{eq:psi}
\psi: \gcal \rightarrow \RR^k, \ \ \psi_l(g) = \PP{L_i = l \cond G_i = g}
\end{equation}
provides a sufficient representation of $G_i$ in the sense of \eqref{eq:repr}:
\begin{equation}
\label{eq:explicit}
\mu(x, \, g) = \frac{\sum_{l = 1}^k  \EE{Y_i \cond X_i = x, \, L_i = l} \PP{X_i = x \cond L_i = l} \psi_l(g)}{\sum_{l = 1}^k \PP{X_i = x \cond L_i = l} \psi_l(g)}.
\end{equation}
\end{lemm}

Expression \eqref{eq:explicit} formalizes the intution laid out in the previous paragraph. The information associated with the category only enters the conditional expectation via the set of probabilities $\PP{L_{i} = \ell \cond G_{i} = g}$. If there are only $k$ latent groups, then each category can be represented in a lossless manner by a $k$ dimensional vector of probabilities. An immediate consequence of this result is that if we knew $\psi$ and gave training examples $((X_i, \, \psi(G_i)), \, Y_i)$ to any universally consistent learner, the learner would eventually recover the optimal prediction function $\mu(\cdot)$. To continue the example at the top of this section, the identity of the hospital enters the model through the probability that a patient is in good or poor health given the hospital.

The dependence of the conditional expectation function $\mu$ on the latent variable probabilities $\psi$ via \eqref{eq:explicit} is non-linear; however, we will retain consistency if we use an expressive enough method for learning on $((X_i, \, \psi(G_i)), \, Y_i)$. Methods known to be universally consistent include $k$-nearest neighbors \citep{stone1977consistent}, various tree-based ensembles \citep{biau2008consistency}, and neural networks \citep{farago1993strong}.

The discussion above seems to imply that we need to estimate $\psi(g) = \PP{ L_{i} | G_{i} = g}$ directly. However, because this quantity depends on the unobservable variable $L_{i}$, its identification is impossible without further assumptions and a more sophisticated approach. Instead we pursue a simpler approach by seeking different functions $f(g)$ that depend only on observables (such as $f(g) = E[X_{i} | G_{i} = g]$), and then proving that they are also sufficient representations because they can be written as invertible functions of $\psi(g)$.

\section{Categorical variable encoding methods}
\label{sec:categorical_encoding}

Our methods proposed below take the form of removing the categorical column and replacing it with a set of columns that can be proven to encode all the categorical information. Each method exploiting the structure mentioned in the previous section. For an overview of other categorical encoding methods already in use, please see section \ref{app:encodings} in the Appendix.

\subsection{Means encoding}
\label{subsec:means}

\begin{figure}[H]
  \centering
  \includegraphics[width=\textwidth]{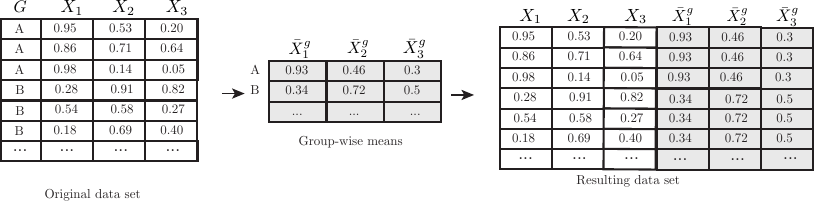}
  \caption{Implementation example of the \emph{means} encoding.\protect\footnotemark}
  \label{fig:means_encoding}
\end{figure}

For our first method, we drop the categorical variables $G_{i}$ and substitute in the average value of the continuous regressors $X_{i}$ given the categorical variable. Figure \ref{fig:means_encoding} shows an illustration.

\begin{figure}[H]
  \centering
  \includegraphics[width=\textwidth]{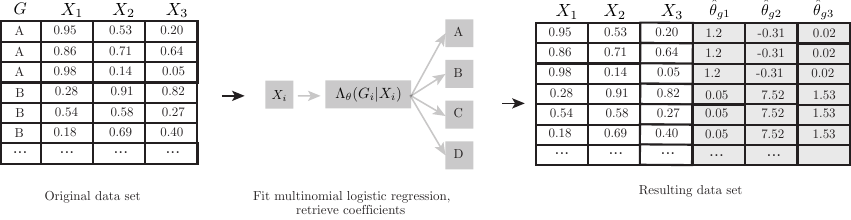}
  \caption{Implementation example of the \emph{mnl} encoding.}
  \label{fig:mnl_encoding}
\end{figure}

\footnotetext{In Figure \ref{fig:means_encoding} and subsequent Figures \ref{fig:lowrank_encoding} and \ref{fig:mnl_encoding}, for easy interpretability, we show the $M x p$ matrix $\hOmega^T$. The reason being that the ultimate implementation of $\psi(g)$ in statistical software requires appending the group encoding to the row the original categorical variable corresponded to.}

\begin{figure}[H]
  \centering
  \includegraphics[width=\textwidth]{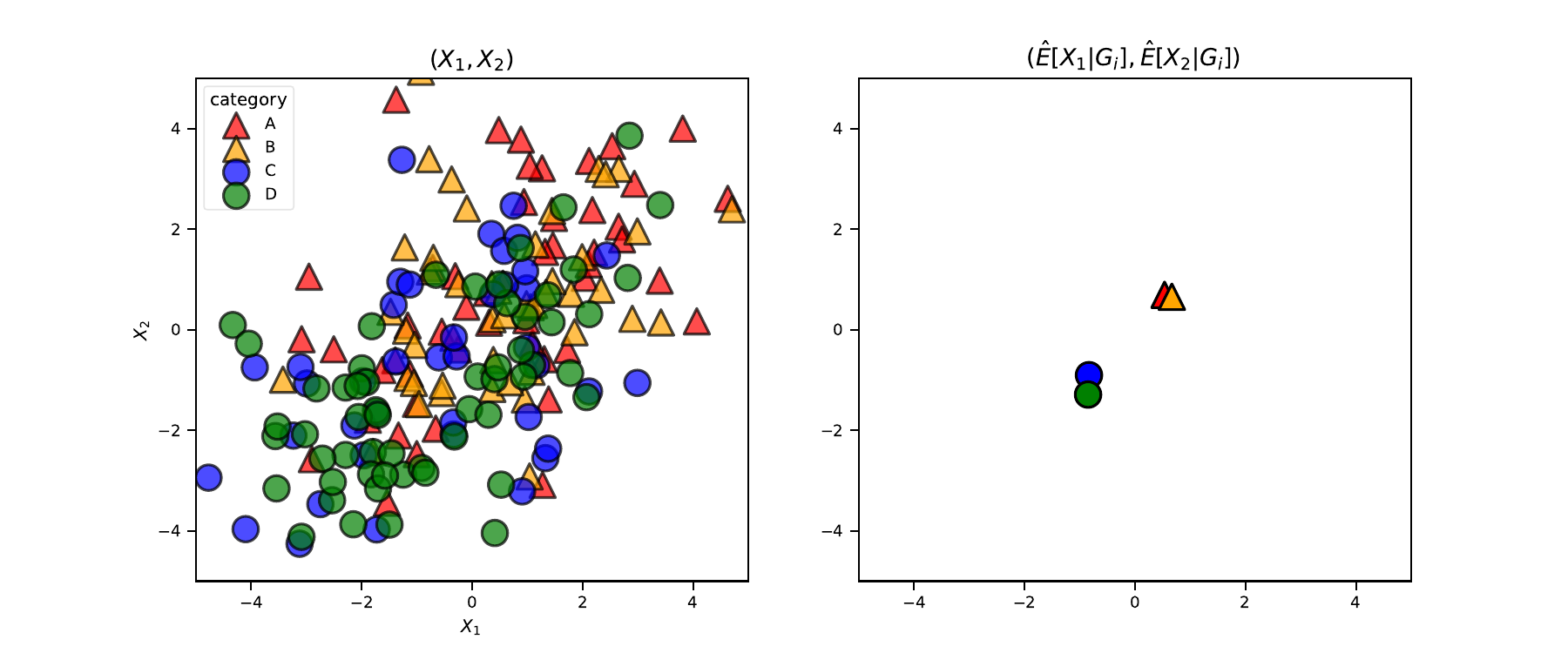}
  \caption{Intution for the \emph{means} encoding on illustrative data. Here, categories $(A,B)$ and $(C,D)$ are associated with separate latent groups.}
  \label{fig:means_intuition}
\end{figure}

This representation is easily interpretable, and it is simple to implement efficiently. This method may be particularly applicable in instances where the number of regressors $p$ is small relative to the number of categories $p \ll |\gcal|$, since then the dimensionality reduction is more dramatic as compared to traditional encoding methods such as one-hot encoding.
Figure \ref{fig:means_intuition} provides an intuitive explanation for why we should expect this to work: the group-wise averages of the continuous variables $(X_{1}, {X_{2}})$ may reveal the dominant latent group in each category.

The next lemma presents the conditions in which this representation provides a sufficient representation. All proofs are in the appendix.

\begin{lemm}
\label{lemm:means}
Under the conditions of Lemma \ref{lemm:repr}, suppose in addition that the matrix $A$ defined~by $(A)_{tj} := \EE{X_{it} \cond L_{i} = l}$
is left-invertible. Then, the $p$-dimensional vectors $\omega(g) := \EE{X_{i} \cond G_i = g}$ are sufficient representations of each category in the sense of \eqref{eq:repr}:
\begin{equation}
\label{eq:explicit_mom}
\mu(x, \, g) = \frac{\sum_{l = 1}^k  \EE{Y_i \cond X_i = x, \, L_i = l} \PP{X_i = x \cond L_i = l} (A^\dagger \omega(g))_l}{\sum_{l = 1}^k \PP{X_i = x \cond L_i = l} (A^\dagger \omega(g))_l}
\end{equation}
\end{lemm}

\begin{algorithm}
\label{alg:means}
\caption{Means Encoding Method}
\begin{algorithmic}[1]

  \Procedure{GroupAverages}{$X,G$}
  \State $\hOmega \gets 0_{p \times M}$
  \Comment{Compute group-wise averages of continuous covariates}
  \For{$g$ in $1$:$M$}

  \vspace{0.09cm}
  \State $\hOmega_{\cdot,g} \gets \frac{1}{|\{i:G_{i}=g\}|}\sum_{i:G_{i}=g}X_{i}$
  \EndFor
  \State \textbf{return} $\hOmega$
  \EndProcedure

  \\

  \Procedure{MeansEncoding}{$X,G$}
  \State $\hOmega \gets$ \textsc{GroupAverages}(X, G)

  \State $S \gets 0_{n \times p}$
  \For{$i$ in $1$:$n$}
  \Comment{Populate with group averages}
  \State $S_{i,\cdot} \gets \hOmega_{\cdot,G_{i}}$
  \EndFor
  \State \textbf{return} $S$
  \EndProcedure

\end{algorithmic}
\end{algorithm}

\subsection{Low-rank encodings}
\label{subsec:lowrank}

The \emph{means} encoding method may efficiently summarize the effect of the categorical variables if the continuous covariates are reasonably low-dimensional so that $p \ll M$. When $p$ is large, it might be beneficial to use a lower-dimensional representation of the conditional means. We suggest two  \emph{low-rank encoding methods}, both involving matrix factorization of the transpose of our group-wise means matrix $\Omega$ of the continuous covariates where $(\Omega)_{jg} = E[X_{ij} | G=g]$.
\begin{figure}[H]
  \centering
  \includegraphics[width=\textwidth]{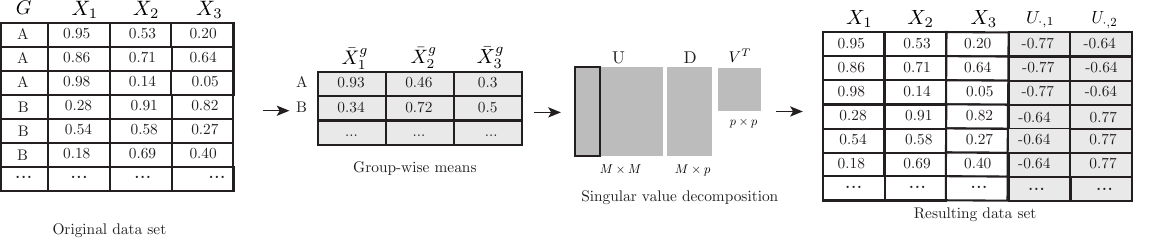}
  \caption{Implementation example of the \emph{low-rank} encoding with singular value decomposition. Alternatively, we could also have used sparse PCA in place of SVD.}
  \label{fig:lowrank_encoding}
\end{figure}

One alternative is to consider the factorization of the transpose of our $p \times M$ group-wise means matrix $\Omega$ using singular value decomposition $\Omega^T = U D V^{T}$. Then we can use the first $k$ columns of the $g^{th}$ row of the left-singular vector matrix $U$ as the representation for the $g^{th}$ category. Note that in practice, we will be working with the empirical counterpart $\hOmega$ and $k$ is in general unknown, so we recommend using cross-validation. Figure \ref{fig:lowrank_encoding} provides an illustration.

A second low-rank alternative is to use sparse principal component analysis (SPCA) method \cite{zou2006sparse} instead of SVD. As the name suggests, this method extends the original PCA algorithm by applying an elastic-net-style penalty on the coefficients of the loadings matrix. The result is that the matrix $\Omega^T$ is approximated by a sparse linear combination of vectors.\footnote{Formally, for a given matrix $M$, the sparse PCA method solves the problem \citep[~eq. 3.12]{zou2006sparse},
\begin{align}
    (\hat{A},\hat{B}) = \arg\min_{A,B} &\sum_{i=1}^n ||M_i - AB^TM_i||^2+ \lambda \sum_{j=1}^k ||B_{\cdot,j} ||_{2}^2 + \sum_{j=1}^k \lambda_{1,j} ||B_{\cdot,j}||_1\\
    \text{s.t.} \quad & A^TA = I_{k \times k}
    \label{eq:sparse_pca}
\end{align}
} This sort of sparsity can be advantageous for two reasons. First, sparse PCA creates principal component sparse vectors which can be potentially more interpretable. Second, if our universal estimator is a tree-based model such as random forest \cite{breiman2001random} or xgboost \cite{chen2016xgboost}, which fit to data by considering singular covariates at any point in the model, it may have difficulty taking advantage of dense principal components due to the rotation of the original covariate space. That is, if a tree would have been able to produce a good split by using each variable separately, it may not be able to do the same if the variables are combined linearly. On the other hand, sparse PCA requires additional tuning of its regularization parameter $\lambda$ which can be done via cross-validation.

Lemma \ref{lemm:lowrank} shows that if indeed there are only $k$ latent groups, then a representation that uses only the first $k$ columns of the left singular matrix or the first $k$ sparse principal components is indeed sufficient. As with the \emph{means} method, we rely on the universal consistency property of our estimator to learn a nonlinear mapping.

\begin{lemm}
\label{lemm:lowrank}
Under the conditions of Lemma \ref{lemm:repr}, suppose in addition that the matrix $A$ defined~by $(A)_{tj} := \EE{X_{it} \cond L_{i} = g}$
is left-invertible. Then, the $k$-dimensional vectors $u(g) := U_{g,1:k}$ for  are sufficient representations of each category in the sense of \eqref{eq:repr}:
\begin{equation}
\label{eq:lowrank}
  \mu(x, \, g) = \frac{\sum_{l = 1}^k  \EE{Y_i \cond X_i = x, \, L_i = l} \PP{X_i = x \cond L_i = l} (A^\dagger V D  u(g)^{T})_l}{\sum_{l = 1}^k \PP{X_i = x \cond L_i = l} (A^\dagger V D  u(g)^{T})_l  }
\end{equation}
\end{lemm}

\begin{algorithm}
\caption{Low Rank Encoding Method}\label{alg:lowrankmethod}
\begin{algorithmic}[1]
\Procedure{LowRankEncoding}{$X,G,k$}

\State $\hOmega \gets$ \textsc{GroupAverages}(X, G)

\State $U,D,V^T \gets SVD(\hOmega^T)$
\Comment{Singular value decomposition}

\State $S \gets 0_{n \times k}$
\For{$i$ in $1$:$n$}
  \Comment{Populate with left singular matrix truncated rows}
  \State $S_{i,\cdot} \gets U_{G_{i},1:k}$
\EndFor
\State \textbf{return} $S$
\EndProcedure
\end{algorithmic}
\end{algorithm}

\begin{algorithm}
\caption{Sparse Low Rank Encoding Method}\label{alg:sparselowrankmethod}
\begin{algorithmic}[1]
\Procedure{SparseLowRankEncoding}{$X,G,k$}

\State $\hOmega \gets$ \textsc{GroupAverages}(X, G)

\State $A, B \gets SPCA(\hOmega^T)$
\Comment{Sparse principal component analysis}

\State $Z \gets \hOmega^T \cdot B_{\cdot,1:k}$
\Comment{Projection on truncated principal components}

\State $S \gets 0_{n \times k}$
\For{$i$ in $1$:$n$}
\Comment{Populate with sparse principal components rows}
\State $S_{i,\cdot} \gets Z_{G_i,\cdot}$
\EndFor

\State \textbf{return} $S$
\EndProcedure
\end{algorithmic}
\end{algorithm}

\subsection{Encoding by multinomial logistic regression coefficients}
\label{subsec:mnl}

Finally, we propose estimating the conditional probability of category membership by multinomial logistic regression parametrized by coefficients $\{\theta_{g} \}_{g \in \mathcal{G}}$
\begin{align}
P(G_{i}|X_{i}) = \Lambda_{\theta}(G_{i}=g | X_{i}) =  \frac{\exp(X_{i}^{T} \theta_{g})}{\sum_{g'}\exp(X_{i}^{T}\theta_{g'})}
\label{eq:mnl}
\end{align}

\noindent and then use the $p$-dimensional vector of coefficients $\theta_{g}$ associated with the $g^{th}$ category to represent it. The motivation for this method comes from the fact that the prediction model $\mu(x, g)$ can be rewritten so that it only depends on the category $g$ through $P(G_{i}=g|X_{i}=x)$, and under the  multinomial logistic regression assumption above this boils down to dependence on the $\theta_{g}$ coefficients.

A related work that also uses regression coefficients as categorical representations is the natural language processing \emph{word2vec} model of \cite{mikolov2013efficient}. The authors of \emph{word2vec} propose two methods to represent words (categories) in a large corpus of text as relatively low-dimensional real-valued vectors. In one of these methods, each word is initially assigned two representations: as a center word $v_{w}$, and as a surrounding \emph{context} word $v_{c}$. Then, the authors posit that the optimal representation is the one that maximizes the log-probability of the inner product of the two representations $v_{w}^{T}v_{c}$ for all pair of words $(w, c)$ that co-occur near each other. Our method works in an analogous way if we let the continuous vectors $X_{i}$ stand in as ``contexts'', and let $\theta_{g} \in \mathbb{R}^{p}$ represent each category, since then maximizing the log-probability of the inner product $X_{i}^{T}\theta_{g}$ is the same as maximizing the multinomial logistic regression above.

\begin{lemm}
\label{lemm:mnl}
Under the conditions of Lemma \ref{lemm:repr}, suppose in addition that $A$ in \eqref{eq:matrix_a} is left-invertible, and that $\PP{ G_{i}=g \cond X_{i}}$ is the multinomial logit distribution with coefficients $\{\theta_{g} \}_{g \in \mathcal{G}}$ containing an intercept. Then, the vector $\theta_{g} \in \mathbb{R}^{p}$ is sufficient in the sense of \eqref{eq:repr}:
\begin{align}
\mu(x, \, g) &= \frac{\sum_{l = 1}^k  \EE{Y_i \cond X_i = x, \, L_i = l} \PP{X_i = x \cond L_i = l} (A^\dagger f(\theta_g))_l}
                     {\sum_{l = 1}^k \PP{X_i = x \cond L_i = l} (A^\dagger f(\theta_{g}))_{l} } \\
&\text{where} \quad f(\theta_{g}) := \frac{\EE[X]{ X_{i}\Lambda_{\theta}(g|X_{i})}}{\EE[X]{\Lambda_{\theta}(g|X_{i})}}
\end{align}
\end{lemm}

\begin{algorithm}
\caption{Multinomial logistic regression method (MNL)} \label{alg:mnl}
\begin{algorithmic}[1]
\Procedure{MNL}{$X, G$}

\State $\htheta \gets \arg\min_{\theta} \sum_{i} \log \Lambda_{\theta}(G_{i}|X_{i})$
\Comment{Multinomial logistic regression}
\State $S \gets 0_{n \times p}$
\For{$i$ in $1$:$n$}
  \Comment{Populate with left singular matrix truncated rows}
  \State $S_{i,\cdot} \gets \htheta_{G_{i}}$
\EndFor
\State \textbf{return} $S$
\EndProcedure
\end{algorithmic}
\end{algorithm}

\section{Experiments}

In the following section, we explore each method's effectiveness relative to one hot encoding across simulated and real world data sets. We apply two typically used methods: random forests and xgboost.\footnote{Simulation code can be found at: \texttt{https://github.com/grf-labs/sufrep}.}

\subsection{Simulations}
\label{sec:simulations}

We consider two simulations designs that share the distributions of latent groups $L_{i}$, observable groups $G_{i}$ and covariates $X_{i}$, but whose outcome models for $Y_{i}$ differ.

\paragraph{Latent groups, observable groups and continuous covariates} A latent group $L_{i}$ is drawn uniformly from the set of available groups, which we identify with integers.
\begin{align}
    L_{i} \sim \text{Uniform}(\{1,...,|\mathcal{L}|\})
    \label{eq:latent_groups}
\end{align}

\noindent Next, observable groups $G_{i}$ are drawn according to the following rule. First, we partition the set of possible observable groups $\mathbb{G}$ into equally-sized sets $\{\mathbb{G}_{\ell}\}_{\ell=1}^{|\mathcal{L}|}$. Then, we draw the observable group $G_{i}$ so that observations that were assigned latent group $L_{1}$ have higher probability of falling into observable group $\mathbb{G}_{1}$, those in $L_{2}$ likely belong to $\mathbb{G}_{2}$, and so on. In symbols,
\begin{align}
    P(G_{i} = g \ | \ L_{i}) =
        \begin{cases}
            \frac{p_{L_{i}}}{|\mathbb{G}_{L_{i}}|}
            \qquad \text{if}\quad g \in \mathbb{G}_{L_{i}}\\
            \frac{1-p_{L_{i}}}{|\mathbb{G}_{L_{i}}^{C}|} \qquad \text{otherwise}
        \end{cases}
    \qquad
    \text{where} \quad p_{L_{i}} > 0.5
\end{align}

Covariates associated with latent group $L_{i} = \ell$ are normally distributed as $X_{i} \sim \mathcal{N}(\mu_{\ell}, \Sigma)$. The mean is zero except for a randomly drawn set of entries $\mathcal{J}$ that are ${-1, +1}$, with $|\mathcal{J}| = 3$.
\begin{align}
  (\mu_{\ell})_{j} =
  \begin{cases}
    0 \qquad &\text{if } j \in \mathcal{J}\\
    \text{Uniform}(\{-1, 1\}) \qquad &\text{otherwise}
  \end{cases}
  \qquad \qquad
  (\Sigma)_{kj} = \left(\frac{1}{2}\right)^{|k - j|}
\end{align}

\paragraph{Outcomes} For the outcome setups, we make each scenario noticeably more complex than the last. In the \emph{global linear} setup, each latent group has its own intercept while the slope $\beta$ is the same across latent groups.
\begin{align}
  Y_{i} &= \alpha_{\ell} + X_{i}^{T}\beta + \epsilon_{i} \label{eq:linear_outcome}
\end{align}
where the intercept and slopes are created as follows. The slope normalization ensures that the signal from the intercept, regressors and noise is roughly comparable.
\begin{align}
\alpha_{\ell} &\sim Laplace(1) \quad \text{for each }\ell \in \mathcal{L} \label{eq:intercept} \\
\tilde{\beta}_{j} &\sim \text{Uniform}(\{0, 1, -1\}) \qquad \beta = \frac{\tilde{\beta}}{\Norm{\tilde{\beta}}_{2}} \label{eq:slopes} \\
\epsilon_{i} &\sim \mathcal{N}(0, 1)
\end{align}

In the \emph{latent linear} setup, we increase the dependence on the latent groups and the outcome model is linear in regressors conditional on coefficients that are specific to each latent group.
\begin{align}
Y_{i} &= \alpha_{\ell} + X_{i}^{T}\beta_{\ell} + \epsilon_{i} \label{eq:latent_outcome}
\end{align}
where
\begin{align}
\tilde{\beta}_{\ell j} &\sim \text{Uniform}(\{0, 1, -1\}) \qquad \beta_{\ell} = \frac{\tilde{\beta_{\ell}}}{\Norm{\tilde{\beta_{\ell}}}_{2}} \label{eq:latentslopes}
\end{align}

Finally, in the \emph{latent piecewise linear} setup, we compute a dyadic basis by partitioning each feature $X_i$ by its median and then assigning different latent group betas ($\beta^+_l$ or $\beta^-_l$) depending on whether or not the observed $X_i$ is above or below its feature's median.
\begin{align}
Y_{i} &= \alpha_{\ell} + \sum_{j=1}^p \textbf{1}\{X_{ij} > \text{Med}(x_j)\} \cdot X_{ij}^T \beta_{\ell j}^{+}  + \textbf{1}\{X_{ij} \leq \text{Med}(x_j)\} \cdot X_{ij}^T \beta_{\ell j}^{-} + \epsilon_{i} \label{eq:nonlinear_outcome}
\end{align}
where
\begin{align}
\tilde{\beta}^+_{\ell j}, \tilde{\beta}^-_{\ell j} &\sim \text{Uniform}(\{0, 1, -1\}) \qquad \beta^{+}_{\ell} = \frac{\tilde{\beta}^{+}_{\ell}}{\Norm{\tilde{\beta}^{+}_{\ell}}_{2}} \qquad \beta^{-}_{\ell} = \frac{\tilde{\beta}^{-}_{\ell}}{\Norm{\tilde{\beta}^{-}_{\ell}}_{2}} \label{eq:nonlinearslopes}
\end{align}



\subsection{Simulation Results}
\label{sec:simulation_results}

For each simulated dataset, we estimated the outcome using the various methods described in Section \ref{sec:categorical_encoding}, and then evaluated the predictions by their mean squared error. We simulated each simulation setup and model for 200 randomly generated seeds.

\begin{figure*}[htbp]
	\centering
	\includegraphics[width=\linewidth]{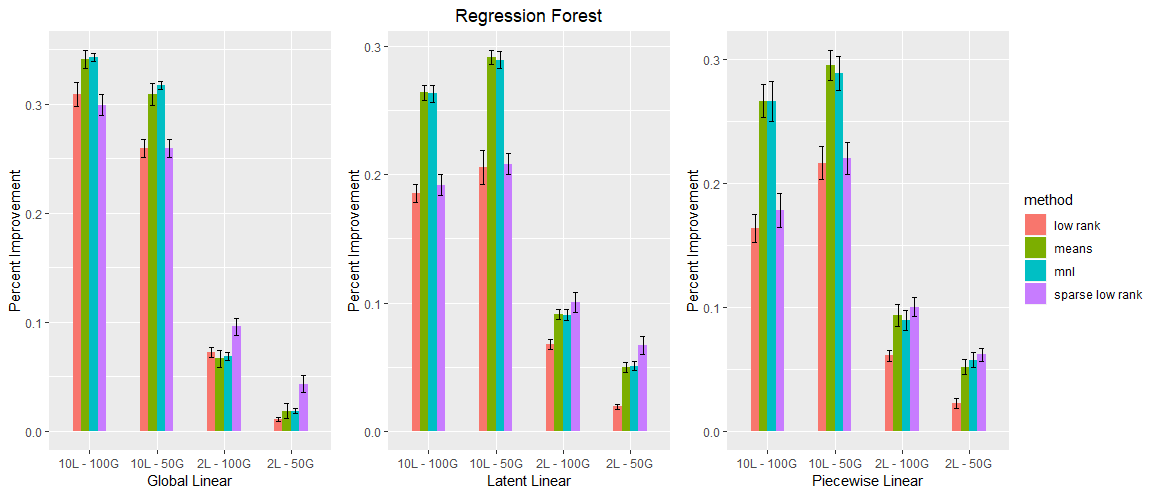}
	\caption {Percent Improvement over One Hot Encoding for Regression Forests.}
	\label{tab:rf_sim_setups}
\end{figure*}

Results for the simulations are provided in \ref{tab:rf_sim_setups}. We find that the methods described above which seek to estimate the latent groups consistently outperform methods which require adding $|\gcal|$ additional columns to our input matrix $X$ for both the regression forest and xgboost. In particular, it appears that the sparse low rank approach tends to do well when the number of latent groups is very small. For a larger number of latent groups, we see that low rank approaches underperform and the multinomial and means encoding perform better. We also take note on how the multinomial weight approach potentially does well in this case possibly because $n$ is large and the number of observations per group is high enough to satisfy this approach.

For the methods that do not take advantage of the low rank structure, we notice that the main improvement in performance for regression forests and xgboost occurs due to the reduction in dimensionality. We find that the permutation, fisher, and multiple permutation methods are on average much better than the methods that add $|\gcal|$ columns but still underperform relative to the methods that estimate the latent groups.

While the performance improvements over one hot encoding for 2 latent groups ranges from 1-10\%, performance improvement can approach 27-33\% for 10 latent groups. Intuitively, we find that this benefit is generally less prevalent for 2 latent groups for regression forests and xgboost due to the lesser complexity of the underlying relationship as defined by the conditional independence graph in \ref{fig:graph_simple}. The improvement over one hot encoding also tends to increase as the signal becomes more dependent on latent group membership. In most cases, performance on the piecewise linear simulations maintain the same or higher percent improvement over one hot encoding. Furthermore, we find that a more complex and nonlinear method like xgboost benefits slightly less from these encoding methods.

\begin{figure*}[htbp]
	\centering
	\includegraphics[width=\linewidth]{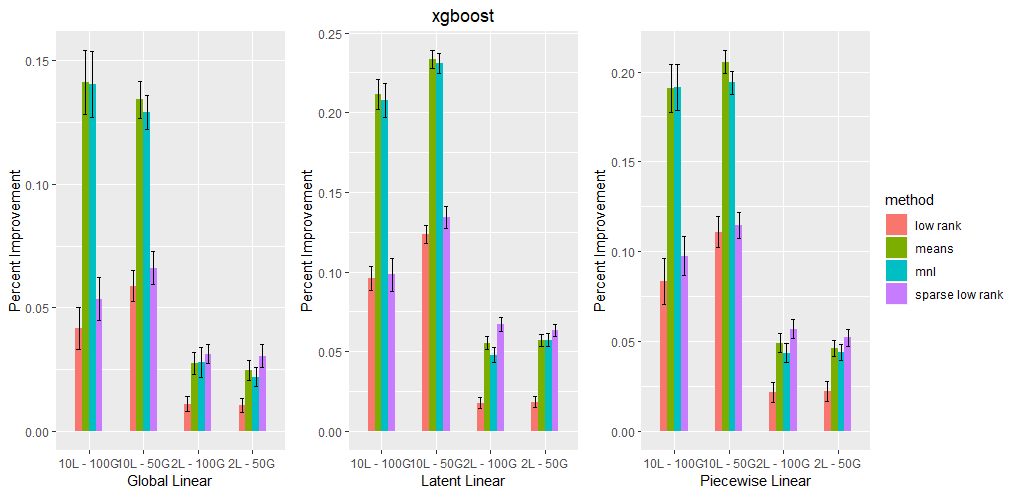}
	\caption {Percent Improvement over One Hot Encoding for xgboost.}
	\label{tab:xgb_sim_setups}
\end{figure*}

\subsection{Empirical Applications}
\label{sec:empirical_applications}

We also evaluate these methods on publicly available datasets that are accessible on Kaggle. We run 4 fold, stratified cross validation on the datasets to avoid the case where there are categorical variables in the test set which are not contained in the training set. Since we are throwing out what could potentially be a sizable amount of information with each fold, we also conduct a paired t-test to further validate or deny results seen in the cross validation process.

\paragraph{Pakistan Educational Performance}

In \cite{pakistanEducation}, Hemani consolidated a series of surveys from Alif Ailaan, a nonprofit organization in Pakistan that focuses on improving education across the country. The objective of the surveys was to provide an objective means of comparing school systems across cities and provinces in order to spark competition between local governments to spur educational reform.

The dataset used in the analysis below contains $n = 580$ and $504$ after removing null valued rows which can be broken down into $|\gcal| = 127$ cities from 2013 to 2016. The number of additional covariates $p$ is 20 and further data cleaning and preprocessing details are elaborated upon in the github.


\paragraph{Ames Housing}

The objective of the Ames Housing Dataset \cite{de2011ames} was to act as a more complex alternative to the Boston Housing Dataset \cite{harrison1978hedonic}. De Cock's aim was to use this dataset for the final project in his regression course which would allow students to more extensively showcase what they had learned.

The Ames Housing Dataset has $n=2,930$ individual home sales in Ames, Iowa from 2006 to 2010. The dataset has 80 covariates and our categorical variable ``neighborhood" has $|\gcal|=25$.

\paragraph{King County House Sales}

The King County House Sales Dataset \citep{houseSalesKingCounty} is a data set containing the record of 21,613 home sales in King County, Washington between May 2014 to May 2015. The author does not provide much additional information aside from it being a good dataset to test regression models. The dataset is relatively popular with over 169,000 views and 28,000 downloads at the time of this paper.

The data itself came with 21 covariates including the sale price of the house. We treat the ``zipcode'' covariate, which has $|\gcal|=70$, as the categorical variable.


\subsection{Empirical Results}

We can see that for Regression Forests on average there is an improvement over one hot encoding and xgboost stands to benefit less from using these encoding methods over one hot encoding. For regression forests, the primary case that does not benefit much from these approaches is the Ames data set which follows naturally since the number of covariates $p$ is much larger than the number of observed groups $|\gcal|$. Therefore, methods such as means and MNL are adding $80$ dimensions to the prediction problem while one hot encoding only adds $25$. The Low Rank and Sparse Low Rank approaches benefit in these cases and appear to maintain potentially promising results. Contrary to the regression forest results, most of the xgboost output was not statistically significantly different than one hot encoding and the Ames data set was the closest evidence to any benefit.

\begin{table}[H]
	\centering
	\begin{tabular}{|c|c|c|c|c|c|}
		\hline
		Dataset& Metric & Means & Low Rank & Sparse Low Rank & MNL  \\
		\hline
		Pakistan & MSE & 9.963 &8.228 &8.868 &8.656 \\
		Pakistan & p-val& 0.00402 & 0.04333&0.00089 &0.01132 \\
		\hline
		Ames & MSE & 1.349 & 1.798&3.987 &-2.120\\
		Ames & p-val&0.73221&0.00930&0.06932&0.81650 \\
		\hline
		Kingcounty & MSE &8.405&8.671&7.062& 8.054\\
		Kingcounty & p-val&0.00445&0.01267&0.03102&0.00364 \\
		\hline
	\end{tabular}
	\caption{Observational Dataset Results for Regression Forests.}
	\label{tab:observational_rf}
\end{table}
For regression forests, on average, it looks like low rank approaches to generating encodings were most robust across data sets. This could be due to the reduction in dimensionality which may be beneficial  for two reasons. First, the underlying relationships were much lower rank than the number of covariates and these methods were able to capture this information. Second, if there was no signal in the categorical variable to begin with, the low rank approaches which utilize K-fold to determine the dimensionality of the encoding are able to pick small $k$ number of encoding vectors to reduce the potential noise covariates one would be adding.


\begin{table}[H]
	\centering
	\begin{tabular}{|c|c|c|c|c|c|}
		\hline
		Dataset& Metric & Means & Low Rank & Sparse Low Rank & MNL  \\
		\hline
		Pakistan & MSE & 2.904 &0.668 & 2.528&-3.391\\
		Pakistan & p-val&0.52714  &0.88127 &0.29955 &0.23304 \\
		\hline
		Ames & MSE & 7.382 & 9.736& 14.889& 1.890\\
		Ames & p-val& 0.31597&0.07341 & 0.13348& 0.59210\\
		\hline
		Kingcounty & MSE & 0.773& -3.243&2.468 & -0.471\\
		Kingcounty & p-val&0.67990&0.62293&0.49389& 0.87640\\
		\hline
	\end{tabular}
	\caption{Observational Dataset Results for XGBoost.}
	\label{tab:observational_xgb}
\end{table}

\section{Application to Doubly Robust Treatment Effect Estimation}
\label{sec:DR}

TODO. We should define the problem with potential outcomes + unconfoundedness; define the DR estimator and briefly summarize its key properties; talk about how our approach to categorical variables fits in; run a simulation + report results.

\section{Conclusion}

In this paper, we explore the task of mapping high-cardinality categorical variables $G_i$ to a lower-dimensional real space without loss of information relevant to our response $Y_i$. To do this, we make an assumption about the relationship between $G_i$ and $Y_i$ which we call the \textit{sufficient latent state assumption}. This assumption provides us with the basis for creating encoding methods which can be used by universally consistent estimators to extract sufficient representations of $G_i$. Among our recommendations for encoding methods, we provide encoding methods which are interpretable or focus more on reducing the size of the  $\mathbb{R}^k$ representation. We find that these methods tend to outperform one hot encoding and other traditional approaches to modeling with categorical variables as the number of unique categories  increases.

\bibliography{references}
\bibliographystyle{plainnat}

\clearpage
\section{Appendix}
\setlength{\parindent}{0pt}

\subsection{Proofs}

\paragraph{Definitions} The following matrices that will be used below.

\begin{align}
    \Omega &=
    \begin{bmatrix}
        \EE{X_{1} \cond G=g_{1}} & \cdots & \EE{X_{1} \cond G=g_{M}} \\
                \vdots     &  \ddots &  \\
        \EE{X_{p} \cond G=g_{1}} & \cdots & \EE{X_{p} \cond G=g_{M}} \\
    \end{bmatrix}_{p\times M}
    \label{eq:matrix_omega}
\\
    A &=
    \begin{bmatrix}
        \EE{X_{1} \cond L=l_{1}} & \cdots & \EE{X_{1} \cond L=l_{K}} \\
                \vdots     &  \ddots &  \\
        \EE{X_{p} \cond L=l_{1}} & \cdots & \EE{X_{p} \cond L=l_{K}} \\
    \end{bmatrix}_{p\times K}
    \label{eq:matrix_a}
\\
    \Psi &=
    \begin{bmatrix}
        \PP{L = l_{1} \cond G=g_{1}} & \cdots & \PP{L = l_{1} \cond G=g_{M}} \\
                \vdots     &  \ddots &  \\
        \PP{L = l_{K} \cond G=g_{1}} & \cdots & \PP{ L = l_{K} \cond G=g_{M}} \\
    \end{bmatrix}_{K\times M}
    \label{eq:matrix_psi}
\end{align}

We denote the columns of $\Omega$ as $\omega(g)$ and the columns of $\Psi$ as $\psi(g)$.

\paragraph{Overview} Proof \ref{proof:suff} shows that categories $G_{i}$ only enter the conditional expectation function $\mu(x, g)$ through the latent state probabilities $\psi(g)$. Proofs \ref{proof:means}-\ref{proof:mnl} rely on strategies for writing $\psi(g) = f(h(g))$ then showing that $h(g)$ is also a sufficient representation.

\subsubsection{Proof of Lemma \ref{lemm:repr}}

\begin{proof} \label{proof:suff}
To show the equivalence of \eqref{eq:mu} and \eqref{eq:explicit}, we begin by expanding \eqref{eq:mu} as
\begin{equation}
\mu(x, \, g)= \sum_{l=1}^L \EE{Y_i \cond X_i = x, \ G_{i} = g, \ L_i = l} \PP{L_i = l \cond X_i = x, \, G_i=g}
\label{eq:mu2}
\end{equation}

Now, the conditional independence assumptions encoded in our graph imply that the expectation term simplifies to
\begin{equation}
\EE{Y_i \cond X_i = x, \ G_{i} = g, \ L_i = l} = \EE{Y_i \cond X_i = x, \ L_i = l}
\end{equation}

while the second term can be rewritten using Bayes rule as
\begin{equation}
\PP{L_i = l \cond X_i = x, \, G_i=g} = \frac{ \PP{X_{i}=x \cond L_{i}=l} \PP{L=l \cond G_{i}=g}  }
       {\sum_{l'=1}^{k} \PP{X_{i}=x \cond L_{i}=l'} \PP{L=l' | G_{i}=g} }
\end{equation}

Combining the above, we see that the mapping $\mu$ only depends on the categorical variable through the multivariable function $\psi(g) = \PP{L_{i}| G_{i}=g}$. Therefore, $\psi(g)$ is a sufficient representation as defined in \eqref{eq:repr}.
\end{proof}

\subsubsection{Proof of Lemma \ref{lemm:means}}

\begin{proof} \label{proof:means}
Begin by noting that conditioanl expectations can be computed as a linear combination of the sufficient statistics discussed in Lemma \ref{lemm:repr}.
\begin{align}
    \EE{ X_{i} \cond G_{i} = g}
    &= \sum_{l=1}^{K} \EE{X_{i} \cond L_{i} = l} \PP{L_{i} = l \cond G_{i} = g} \\
    &= \sum_{l=1}^{K} \EE{X_{i} \cond L_{i} = l} \psi_{l}(g)
    \label{eq:scalar_decomposition}
\end{align}

\noindent or, in matrix form,
\begin{align}
    \Omega = A\Psi
    \label{eq:matrix_decomposition}
\end{align}

where these matrices are defined as in the top of this section. The sufficient representation for the category $\psi(g) = \Psi_{g}$ lies on the linear span of the set of columns of $A$. Since $A$ has a left-inverse $A^\dagger$ such that $A^\dagger A = I$, we can retrieve the representations by matrix multiplication.
\begin{align}
    \psi(g) = (\Psi)_{\cdot, g} = A^\dagger(\Omega)_{\cdot, g} =: A^\dagger \omega(g)
    \label{eq:omega_inverse}
\end{align}

Since $\psi(g)$ only depends on $g$ through $\omega(g)$, it follows that $\omega(g)$ is also a sufficient representation for the category $g$.
\end{proof}

\subsubsection{Proof of Lemma \ref{lemm:lowrank}}

\begin{proof} \label{proof:lowrank}
  The proof is similar to the the previous one. However, this time note that we can decompose the $\Omega^T = U D V^{T}$ using singular value decomposition, where the matrices have dimensions $|\mathcal{G}| \times |\mathcal{G}|$, $|G|\times p$, and $p \times p$ respectively. Letting $u(g) : g \mapsto (U)_{g, 1:k}$, we can write
  \begin{align}
    \psi(g) = A^\dagger V D^T u(g)^T
  \end{align}
  where $D$ and $V$ do not depend on the category $g$ and $k$ is the true number of latent groups and column rank of $\Omega^T$. Finally, to complete the proof, we substitute $V D u(g)^T$ with $\omega(g)$ in \eqref{eq:omega_inverse}.


\end{proof}

\subsubsection{Proof of Lemma \ref{lemm:mnl}} \label{proof:mnl}
\begin{proof}
We begin by noting that we can use Bayes' theorem to express $\omega(g) = \EE{ X_{i} \cond G=g }$ as a function of $\PP{G=g \cond X_{i} }$, here is assumed to be multinomial logit.

\begin{align}
  \EE[X|G] { X_{i} \cond G_{i} = g }
  &=  \EE[X] { X_{i} \PP{X_{i} | G_{i} = g} }  \\
  &= \frac{\EE[X]{ X_{i} \PP{G_{i} = g|X_{i}}}}{\PP{G_{i} = g}} \label{eq:mnl_bayes} \\
  &= \frac{\EE[X]{ X_{i} \PP{G_{i} = g|X_{i}}}}{\EE[X] { \PP{G_{i} = g \cond X_{i}}}}  \\
  &= \frac{\EE[X]{ X_{i} \Lambda_{\theta}(g|X_{i}} }{ \EE[X]{\Lambda_{\theta}( g | X_{i})}}  \label{eq:mnl_logit}
\end{align}

However, note that expression \eqref{eq:mnl_logit} only depends on the category through the mutinomial logit coefficients $\theta_{g}$ that are associate with category $g$. Therefore, under this assumption we can write $\omega(g) = f(\theta_g) =: E[X_{i} | G=g]$. However, recall from \eqref{eq:omega_inverse} that if the matrix $(A)_{j\ell} := E[X_{ij} | L_{i}=\ell]$ has a left-inverse $A^{\dagger}A = I$, we can write
\begin{align}
\psi(g) = A^\dagger \omega(g) = A^\dagger f(\theta_{g})
\end{align}

Since $\psi(g)$ only depends on $g$ through $\theta(g)$, it follows that $\theta(g)$ is also a sufficient representation for the category $g$.
\end{proof}

\subsection{Additional Encoding Methods}{\label{app:encodings}}

For a more in-depth treatment, see \cite{venables2016codingmatrices}. Note that several of the methods below are simple linear transformations of each other and should yield equivalent levels of performance in theory. However, as we will see in sections \ref{sec:simulation_results} and \ref{sec:empirical_applications}, in practice the resulting performance can differ substantially.

\paragraph{One-hot or dummy} This is the most common categorical encoding, and it is the method we take to be our main baseline, against which we will compare all other methods. It expands out the categorical column into $k-1$ columns where $k$ is the number of unique elements in the set of categorical levels in the column. Each column is binary 1 or 0 depending on whether the corresponding level was observed in the original categorical column. \citep[~sec 2.3.2]{murphy2012machine}

\begin{table}[H]
	\centering
	\begin{tabular}{rrrrr}
		\hline
		& b & c & d & e \\
		\hline
		a & 0 & 0 & 0 & 0 \\
		b & 1 & 0 & 0 & 0 \\
		c & 0 & 1 & 0 & 0 \\
		d & 0 & 0 & 1 & 0 \\
		e & 0 & 0 & 0 & 1 \\
		\hline
	\end{tabular}
\end{table}

\paragraph{Deviation} Similar to one-hot encoding except that the $k^{th}$ unique element's row that is the reference level is now set to all values of $-1$. This means that categorical levels are being compared to the grand mean of all of the levels instead of the mean of a given level with respect to the reference level.

\begin{table}[H]
	\centering
	\begin{tabular}{rrrrr}
		\hline
		& b & c & d & e \\
		\hline
		a & 1 & 0 & 0 & 0 \\
		b & 0 & 1 & 0 & 0 \\
		c & 0 & 0 & 1 & 0 \\
		d & 0 & 0 & 0 & 1 \\
		e & -1 & -1 & -1 & -1 \\
		\hline
	\end{tabular}
\end{table}

\paragraph{Difference} Compares a given level to the mean of the levels that precede it.

\begin{table}[H]
	\centering
	\begin{tabular}{rrrrr}
		\hline
		& b & c & d & e \\
		\hline
		a & -0.5 &-0.333& -0.25& -0.2\\
		b &  0.5 &-0.333& -0.25& -0.2\\
		c &  0.0 & 0.667& -0.25& -0.2\\
		d & 0.0 & 0.000 & 0.75 &-0.2\\
		e &  0.0&  0.000&  0.00&  0.8\\
		\hline
	\end{tabular}
\end{table}

\paragraph{Helmert} Compares levels of a chosen categorical variable to the mean of the subsequent levels uniquely observed thus far.

\begin{table}[H]
	\centering
	\begin{tabular}{rrrrr}
		\hline
		& b & c & d & e \\
		\hline
		a & 0.80 & 0.00 & 0.00 & 0.00 \\
		b & -0.20 & 0.75 & 0.00 & 0.00 \\
		c & -0.20 & -0.25 & 0.67 & 0.00 \\
		d & -0.20 & -0.25 & -0.33 & 0.50 \\
		e & -0.20 & -0.25 & -0.33 & -0.50 \\
		\hline
	\end{tabular}
\end{table}

\paragraph{Repeated Effect} Columns are encoded to represent a cumulative comparison of subsequent levels with previous ones.

\begin{table}[H]
	\centering
	\begin{tabular}{rrrrr}
		\hline
		& b & c & d & e \\
		\hline
		a & 0.8 & 0.6 & 0.4 & 0.2\\
		b & -0.2 & 0.6 & 0.4&  0.2\\
		c & -0.2& -0.4 & 0.4 & 0.2\\
		d & -0.2 &-0.4& -0.6&  0.2\\
		e & -0.2& -0.4& -0.6 &-0.8\\
		\hline
	\end{tabular}
\end{table}

\paragraph{Permutation} Assigns a unique integer to each category. Note that even when the categories do not possess an intrinsic ordering, some mappings may yield better results if they happen to be aligned with the true average effect the category has on the outcome variable.

\begin{table}[H]
	\centering
	\begin{tabular}{rrrrr}
		\hline
		& perm  \\
		\hline
		a & 5  \\
		b & 3  \\
		c & 4  \\
		d & 1 \\
		e & 2  \\
		\hline
	\end{tabular}
\end{table}

\paragraph{Multi-Permutation (Multi-Perm)} Following the intuition above, with a larger number of columns we might find more interesting permutations. Hence, we also experiment with four random integer mappings at once.

\begin{table}[H]
	\centering
	\begin{tabular}{rrrrr}
		\hline
		& perm1 & perm2 & perm3 & perm4 \\
		\hline
		a & 1 & 5 & 4 & 2 \\
		b & 2 & 3 & 5 & 3 \\
		c & 3 & 1 & 2 & 4 \\
		d & 4 & 4 & 1 & 1 \\
		e & 5 & 2 & 3 & 5 \\
		\hline
	\end{tabular}
\end{table}

\paragraph{Fisher} taken from \cite{hastie2009elements}, we order the categories by increasing mean of the response.

\vspace{.5em}

For the following five methods, we use information about the continuous covariates to construct the mapping $\psi$.

\end{document}